\documentclass{aa}

\usepackage{graphicx}
\usepackage{graphics}
\usepackage{amsmath}
\usepackage{float}
\usepackage{caption}
\usepackage{subcaption}

\usepackage{txfonts}
\usepackage{soul,color}
\begin{document}

   \title{Convolutional Neural Networks applied to sky images for short-term solar irradiance forecasting}

   %\subtitle{EU-PVSEC}

   \author{Q. Paletta
          \inst{1}
          \and
          J. Lasenby
          %C. Ptolemy\inst{2}\fnmsep\thanks{Just to show the usage
          %of the elements in the author field}
          \inst{2}
          }

   \institute{University of Cambridge, Engineering Department\\
              %T\"urkenschanzstrasse 17, A-1180 Vienna\\
              \email{qp208@cam.ac.uk}
              %\thanks{+33(0)671912858}
         \and
             University of Cambridge, Engineering Department\\
             \email{jl221@cam.ac.uk}
        %     \thanks{The university of heaven temporarily does not
        %             accept e-mails}
             }

%   \date{Received September 15, 1996; accepted March 16, 1997}

% \abstract{}{}{}{}{} 
% 5 {} token are mandatory
 
  \abstract
   {Reducing carbon emissions is critical to meeting international targets aimed at mitigating climate change. One way to reduce emissions is for countries to decarbonise their energy production in the upcoming decades. For this reason, the share  of renewables is expected to increase in the global energy mix. In particular, the importance of solar energy for electricity production is set to rise. However, despite seamless advances in the estimation of the solar resource, there is still a need for better solar forecasting to improve its integration into the energy supply. Fish-eye cameras are emerging in-situ meteorological sensors that have already demonstrated promising and interesting results for high temporal resolution and very short term solar forecasting. The many innovations in the fields of machine learning and computer vision are likely to improve upon the performance of current solar forecasting techniques; this is to be expected given the complexity of the sky and cloud movement when viewed in high resolution. This work presents preliminary results of the use of deep Convolutional Neural Networks (CNNs) for short-term solar irradiance forecasting using sky images.
   \vspace{0.5\baselineskip}
}
  % aim and approach heading (mandatory)
   {To facilitate its integration and increase its economic value, solar energy would benefit from improvements in the forecasting of electricity generation at short (10 minutes), medium (1 hour) and long (1 day) term timescales. Tools available range from satellite imagery analysis and statistical modelling to ground-based sky image analysis: the latter provides very short term information on cloud cover changes which can be used, for example, to predict clouds hiding the sun. However, current approaches to model the cloud cover dynamics from sky images still lack precision regarding the spatial configuration of clouds, their temporal dynamics and their physical interaction with solar radiation.
   
   \vspace{0.5\baselineskip}
   The approach introduced by~\cite{zhang2018}, \cite{Siddiqui2019}, \cite{zhao2019}, which we follow here, is to apply recent computer vision techniques from the field of Deep Learning to forecast solar radiation up to 20 minutes in advance, from a range of data including sequences of past ground-based sky images. Artificial neural networks are implemented to extract relevant features from a dataset of around 64,000 images and in-situ pyranometric measurements taken from February to September 2018 in Palaiseau with a temporal resolution of 2-min \citep{sirta}.
   \vspace{0.5\baselineskip}
   }
  % scientific innovation and relevance heading (mandatory)
   {This work aims at bringing innovative insights via a novel approach to irradiance forecasting using the Deep Learning framework, which constitutes an effective environment for a richer modelling of the cloud cover and its dynamics. The model performance has been evaluated using the Mean Square Error (MSE) as error metric and a persistence of the clear-sky index as a reference to calculate the corresponding skill score.
    Also, visualisation methods were implemented to understand what the model has learnt during training and what region of the image it is using to make its predictions.
    
   \vspace{0.5\baselineskip}}
   {The study shows that CNNs are able to successfully estimate future irradiance from a sequence of past images of the sky. The corresponding 10-min forecast skill based on the Mean Square Error (MSE) reaches 40\% when evaluated on a set of 4000 unseen pairs of images. In practice, samples from the training and validation sets were drawn randomly from two groups comprising data from distinct days. A secondary approach involving training the model with past data from the same day showed a 10\% performance improvement on the skill score compared to the previous setting, which outlines the need to incorporate historical data of the day in short term forecasting.
   
   \vspace{0.5\baselineskip}
   A complementary investigation was also conducted to understand what relevant features the Deep Learning model is extracting from sky images. In particular, images maximising the response to each filter of the CNN are generated to visualise the different patterns used by the model.
   \vspace{0.5\baselineskip}}
   {The results obtained indicate that, despite a tendency to mimic the persistence model, the Deep Learning model spots relevant patterns in sky images to adapt its forecast. It is not yet able to fully anticipate sudden irradiance changes from a sequence of images, but it would benefit from incorporating historical data in its predictions. The current architecture of the network could be further upgraded with recurrent units to facilitate the extraction of relevant features from the temporal aspect of the data.
   \vspace{4\baselineskip}
   }


\maketitle
%\vspace{2\baselineskip}
 %Distinct days & 0.20 & 0.40 & \textbf{0.43} \\

  %\titlerunning 
%
%-------------------------------------------------------------------

\newpage
\pagebreak

\section{Contribution}

The proposed approach introduced by \citet{zhang2018}, \citet{Siddiqui2019} and \citet{zhao2019} is to apply Deep Learning models to extract patterns not only from single point data but also from sky images. In particular, 2D inputs such as images can be treated by a specific type of neural network termed a CNN. The learning of such models is based on the use of filters which are trained to find task relevant patterns in an image. In parallel with other past meteorological data, such information can be given to an Artificial Neural Network (ANN) to forecast irradiance. The global network made of the CNN and the ANN can be trained simultaneously using supervised learning. The original contributions of this study are the following:
\vspace{-0.5\baselineskip}

\begin{itemize}
    \item \citet{Siddiqui2019} uses CNNs to extract patterns on sky images from the past 4 to 6 hours as general information about the cloud cover of that day to adapt forecasts up to 6h ahead. We, however, set up our protocol to learn how the clouds visible from the lens at time $t$ will induce the variability of the solar resource in the short-term upcoming future. In other words, we limit the forecasting time window to a narrower 20-min window from auxiliary data (past irradiance measurements, solar position of the sun and its sine / cosine transformations) and images taken at time $t$ and $(t-2)$min.
    
    \vspace{0.4\baselineskip}
    
    \item Instead of training and validating the model on distinct days, which does not let the model learn from same day past samples, we tried a new setting to generate both sets, which improves forecasting performances by around 10\%. With this approach, the model is validated on afternoon samples only, morning samples being in the training set. This way, the protocol is closer to real life applications where past samples of the same day are available for model fine-tuning.
    
    \vspace{0.4\baselineskip}
    
    \item To assess the model's performance we implemented the MSE forecast skill based on the smart persistence model, which is using the last in-situ pyranometric measurements only. A clear-sky index is computed with a concomitant model of clear-sky (here the ESRA model). This clear-sky index is then multiplied by the corresponding clear-sky prediction of the model to obtain the smart persistence forecast. This gives a reliable performance score to compare the proposed Deep Learning model with other approaches.
    
    \vspace{0.4\baselineskip}
    
    \item To increase the receptive field of the network without using dilated convolution \citep{Dilated_convolutions,Siddiqui2019}, we implemented a ResNet architecture \citep{Resnet}, which enables solvers to handle a deeper model.
\end{itemize}

\section{Approach}

\subsection{Dataset}
\label{dataset}
%SIRTA, training and validation set \\

\begin{figure}[H]
\centering
\begin{subfigure}{0.23\textwidth}
    \centering
    \includegraphics[width=1\linewidth]{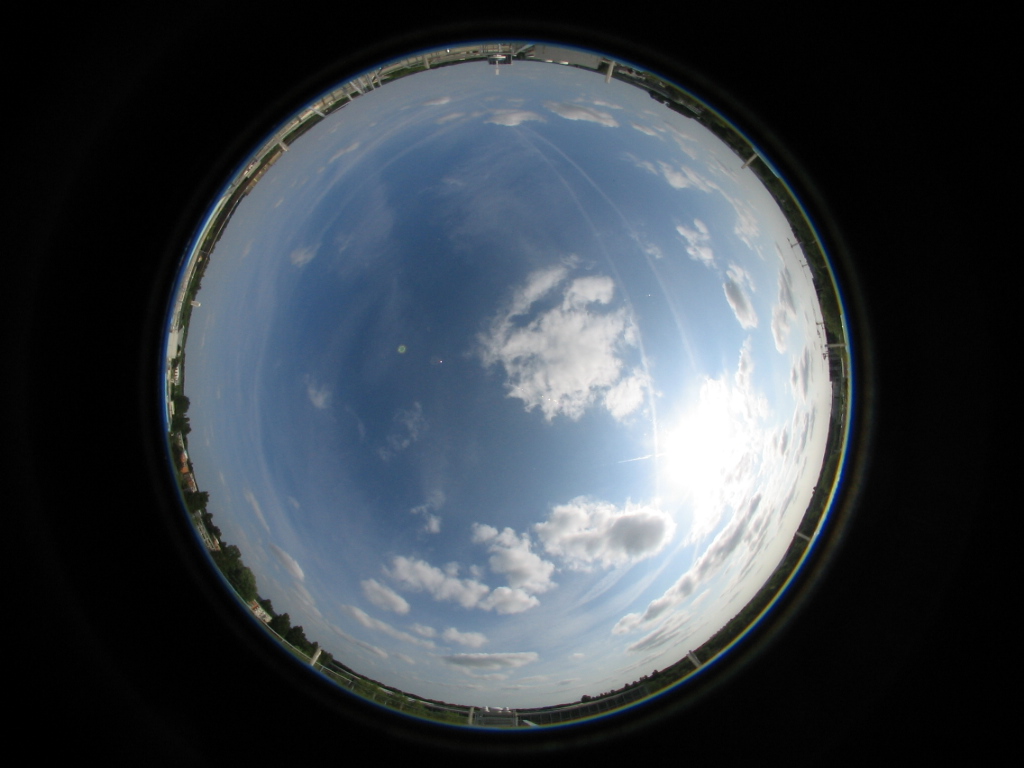}
    %\caption{Long exposure (source:~\cite{sirta})}
    %\label{fig:sub2}
\end{subfigure}
\begin{subfigure}{0.23\textwidth}
    \centering
    \includegraphics[width=1\linewidth]{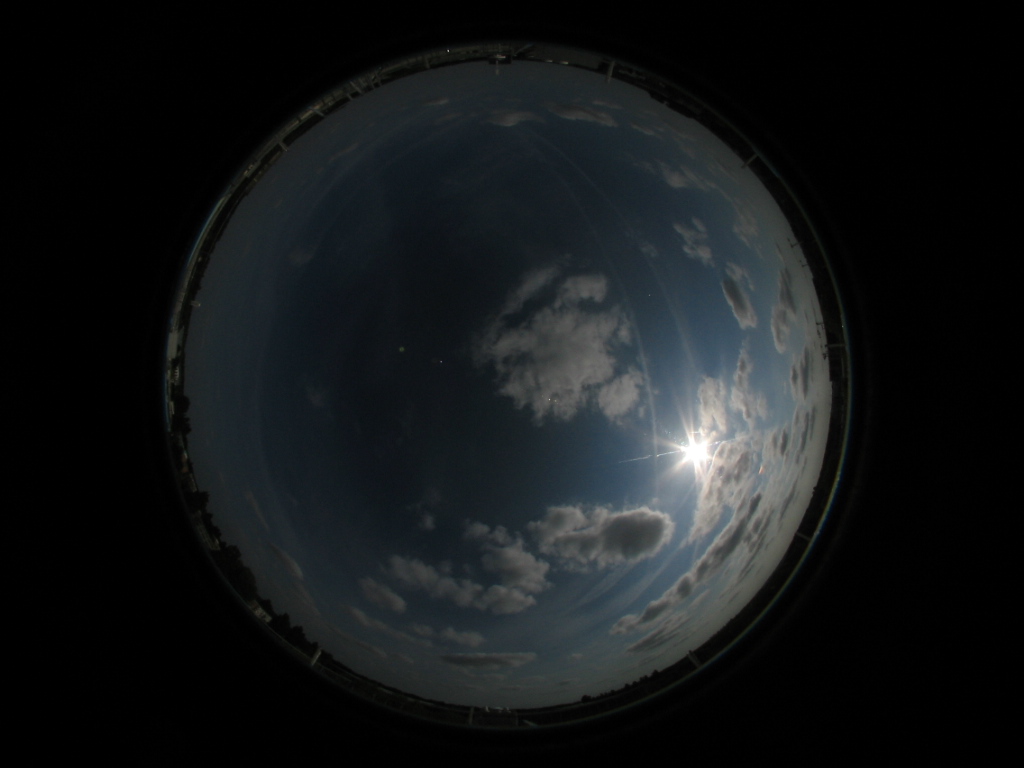}
    %\caption{Short exposure (source:~\cite{sirta})}
    %\label{fig:sub1}
\end{subfigure}%
\caption{Images of the sky taken by a hemispherical camera with short and long exposures (source:~\cite{sirta})}
\label{fig:img_short_long}
\end{figure}

The chosen approach is to provide the Deep Learning model with sky images taken by hemispherical cameras on the ground (see Figure~\ref{fig:img_short_long}) and a range of metadata such as past irradiance measurements or the angular position of the sun.

%\vspace{0.5\baselineskip}

\vspace{0.5\baselineskip}
The dataset used in this study originated from the SIRTA laboratory \cite{sirta}. Samples were collected over a period of seven months from March 2018 to September 2018. Images of the sky have RGB resolution of 768 by 1024 pixels. Samples are taken every two minutes and composed of two images with different exposures (referred to as long and short exposures, see Figure~\ref{fig:img_short_long}).

\vspace{-0.7\baselineskip}

The short exposition time provides more details on the region close to the sun, whereas the long exposure focuses on the rest of the sky, and in particular on distant clouds. In addition to this, the dataset has a range of metadata, in particular the Global Horizontal Irradiance (GHI) measurements which are averaged over a minute. Also, the angular position of the sun is available through the Zenith and Azimuthal angles.

\subsection{Network Architecture}
\label{network_architecture}

\begin{figure}[H]%[ht!]%[h!] 
\centering    
\includegraphics[width=0.5\textwidth]{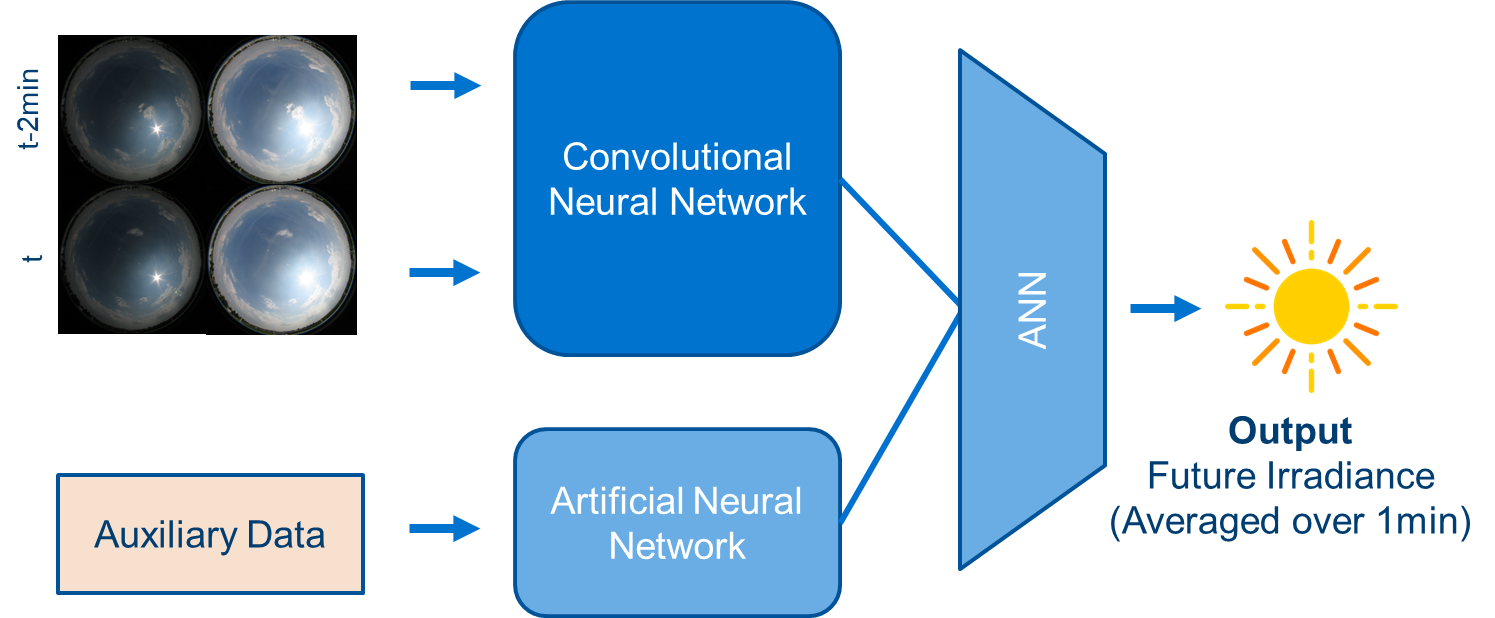}
\caption{Schematic network architecture}
\label{fig:network_architecture_2}
\end{figure}

The proposed model follows the configuration presented in Figure~\ref{fig:network_architecture_2}. It is composed of two distinct networks merged into one which output the irradiance estimate. On the one side, a CNN made of ResNet units is used to extract features from sky images and on the other side, an ANN treats available auxiliary data (past irradiance measurements, angular position of the sun, etc). Both outputs are fed into another ANN, which integrates them to give its prediction.

%\hl{1 Choix de l'architecture originale}
%\hl{2 adaptation pour ton problème}
%\hl{3 hyperparamètres}

\vspace{-0.5\baselineskip}
Hyperparameters  of the model such as the number of filters / nodes per layer or the number of layers are mostly hand-tuned given the model's performance. Another approach presented in Section~\ref{models_leaning} was used to select the number of filters per layer. In addition, the number of layers, and in particular the number of convolutional layers with a stride of 2 (the filter convolves around the input volume by shifting two units at a time) was set such that the output of the last convolutional layer was computed from each pixel of the input images.

\begin{figure}[H]%[ht!]%[h!] 
\centering    
\includegraphics[width=0.34\textwidth]{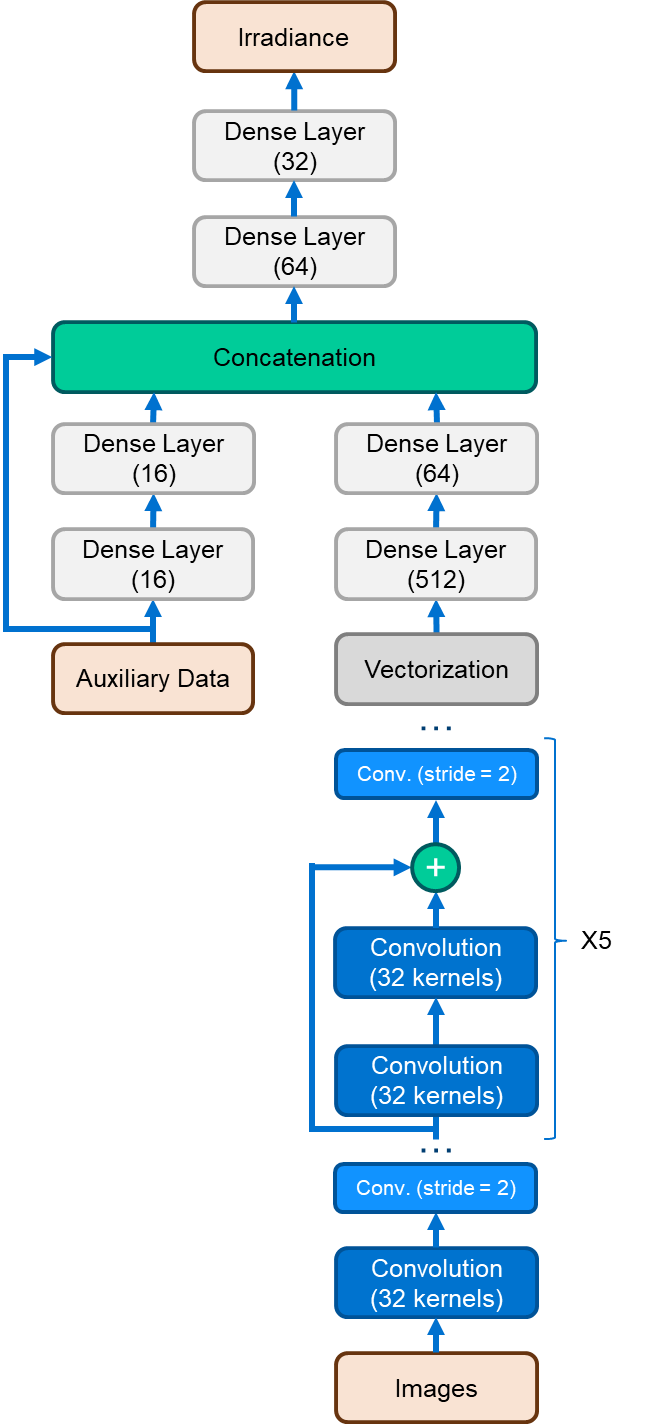}
\caption{Detailed network architecture}
\label{fig:network_architecture_1}
\end{figure}

\vspace{0.5\baselineskip}
The CNN architecture is composed of a set of convolutional layers with 32 filters in each and strides of 1 or 2. ResNet units using residual connections are implemented to increase the depth of the network. This way, if a set of convolutional layers does not improve the overall performance, the network is able to bypass the given 'residual connection' through a parallel identity connection (see Figure~\ref{fig:network_architecture_1}). The vectorisation layer reshapes the 2D output of the last convolutional layer into a 1D layer whose nodes are connected to dense layers with 512 and 64 nodes.

\vspace{0.5\baselineskip}
The parallel ANN is composed of two layers of 16 nodes plus a residual connection, which merges with the other network through a concatenation layer. Following this, another set of 2 densely connected layers integrate the incoming information to output the irradiance estimate through the final node.

\subsection{Training and Validation}
\label{training_and_validation}

%Also the pixel to pixel difference between two time frame is send to the model (see Figure~\ref{})

Each sample given to the network is composed of four images in total, the short and the long exposure taken at time $t$ and $(t-2)$min. Each 2D input was originally downsized to a grey scale 150x150px resolution. In parallel, metadata are made of the past irradiance measurements at time $t$ and $(t-2)$min, the angular position of the sun at time $t$ defined by its Zenith and Azimuthal angles, plus the cosine and sine of those angles.

\vspace{0.5\baselineskip}
%To highlight the point of this study, t
The assessment of the model was performed on a setting of the training and the validation sets composed of 16,000 and 4000 samples respectively. Samples collected from 8am to 7pm over the seven months period were randomly allocated to the validation and the training sets from two groups comprising data from distinct days. This way we prevent the model from using samples of the same day seen in the training set to adapt its forecasts on samples of the validation set.

\vspace{0.5\baselineskip}
In a second setting however, we tried to shift morning samples from the validation set to the training set to see how training the model on past samples of the same day improves its predictions. This configuration could be seen as closer to real life applications as data are streamed continuously and could be used to fine-tune the model given previous measurements.

\vspace{0.5\baselineskip}
The loss function used by the model as a reference to assess its own performance is the unregularised MSE. The learning rate associated with the minimisation of the loss was set to $10^{-3}$. The performance of the model was assessed using the forecast skill metrics based on the persistence of the clear-sky index on both training and validation set settings.

\vspace{0.5\baselineskip}
Hyperparameter tuning was performed to achieve the best forecasting performance on the 10-min ahead forecast. The same network architecture has been used to train models for the 2-min to 20-min ahead forecasts. The only difference is that the number of filters per convolutional layer was reduced to 16 for the 2-min and 4-min ahead forecasts due to a tendency to overfitting with 32 filters.

%\begin{figure}[H]%[ht!]%[h!] 
%\centering    
%\includegraphics[width=0.4\textwidth]{Figs/fs_validations_2.png}
%\caption{Comparison of both settings using the MSE Forecast Skill based on the smart persistence model on the validation set}
%\label{fig:fs_validations_2}
%\end{figure}

%0.21 0.37 0.43

%The results obtained for different input settings are presented in table~\ref{table:results}. One can see that integrating both auxiliary data and sky images improves the performance of the model, especially for the 10-min ahead forecast, which was chosen as the reference forecast to select a shared set of hyperparameters for the other models. Therefore, finding a specific set of hyperparameters for each forecast window could improve individual performances.

%\vspace{0.5\baselineskip}

\section{Results}
\label{results}

\begin{figure}[H]%[ht!]%[h!] 
\centering    
\includegraphics[width=0.38\textwidth]{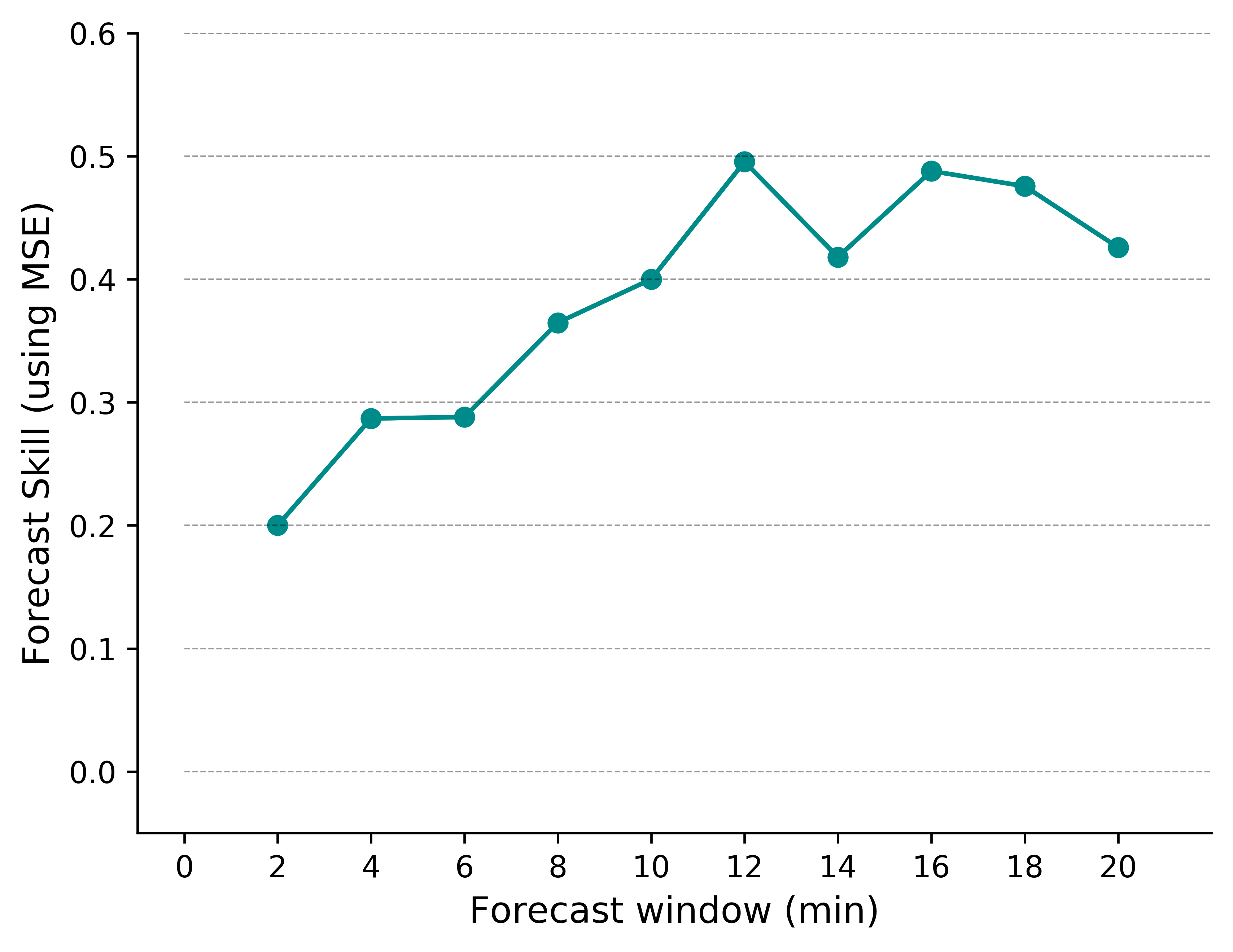}
\caption{Forecasting skill on the validation set using the MSE based on the smart persistence model for different time windows, the training and validation sets being generated from different days}
\label{fig:fs_different_time_windows}
\end{figure}

Figure~\ref{fig:fs_different_time_windows} shows the performance of the model obtained with the same architecture for different time windows from 2 to 20-min. Contrary to~\cite{zhang2018}, the longer the forecast window, the better the forecast skill, until it reaches a plateau from the 12-min forecast window, which tempers the conclusion of~\cite{zhang2018} about a decrease of the skill score for longer time horizon forecasts.

\vspace{-0.5\baselineskip}

%Also, the authors would like to mention that including training samples of the 

%\vspace{-0.5\baselineskip}

Also, it is interesting to note that showing past data of the same day during the training strongly increases the model's performance with a skill score on the 10-min ahead forecast increasing by around 10\% from 0.4 to 0.44 for instance (see Table~\ref{table:results}). These results highlight the need to take same day samples of the past into account even for short term forecasting. This could be achieved through active learning or forecasting from a longer sequence of samples from the past (see~\cite{Siddiqui2019}).

\renewcommand{\arraystretch}{1.1}
\begin{table}[H]
%\vskip 0.15in
\begin{center}
\begin{small}
%\begin{sc}
\begin{tabular}{|c|ccc|}
\hline
Settings & 2-min & 10-min & 20-min\\
%\cline{2-4}
%& MSE & RMSE & MAE \\
\hline
Distinct days & 0.20 & 0.4 & 0.43 \\
Validation on afternoon samples & \textbf{0.26} & \textbf{0.44} & \textbf{0.45} \\
\hline
\end{tabular}
\end{small}
\end{center}
\caption{Forecast Skill (MSE) on the validation set based on the smart persistence model}
\label{table:results}
\vskip -0.1in
\end{table}

\section{Model's learning assessment}
\label{models_leaning}

Often considered as black boxes, Deep Learning models foster research focusing on their interpretability. In that respect, different methods have been proposed to visualise models' learning, from intermediate activations and filter visualisation \citep{dl_with_python} to feature maps \citep{Zeiler2013}.

\vspace{-0.7\baselineskip}

\subsection{Intermediate activation visualisation}

\vspace{-0.8\baselineskip}

\begin{figure}[h!]
    \begin{minipage}[b]{0.55\linewidth}
        Visualising the layer activation of a trained model is useful for two reasons. Firstly, one can see what trained filters focus on for a given input. Figure~\ref{activation_visualisation} shows the transformation of figure~\ref{activation_visualisation_image}, once passed through 60 filters of the first convolutional layer. As we can see, some filters focus on specific patterns of the image such as the sun or the distant sky. However, some filters seem to look at useless areas of the image such as the black background or even simply return a blank output.

        %\vspace{2.5\baselineskip}
    \end{minipage}\hfill
    \begin{minipage}[b]{0.41\linewidth}
        \centering
        \includegraphics[width=0.9\textwidth]{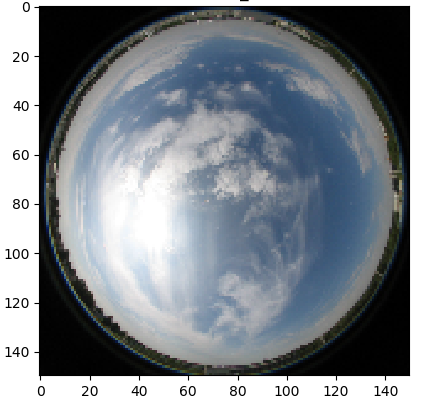}
        \caption{Example of an image of the sky, which is passed through the first layers to visualise intermediate activations (see Figure~\ref{activation_visualisation})}
        \label{activation_visualisation_image}
    \end{minipage}
\end{figure}

\begin{figure}[H]%[ht!]%[h!] 
\centering    
\includegraphics[width=0.46\textwidth]{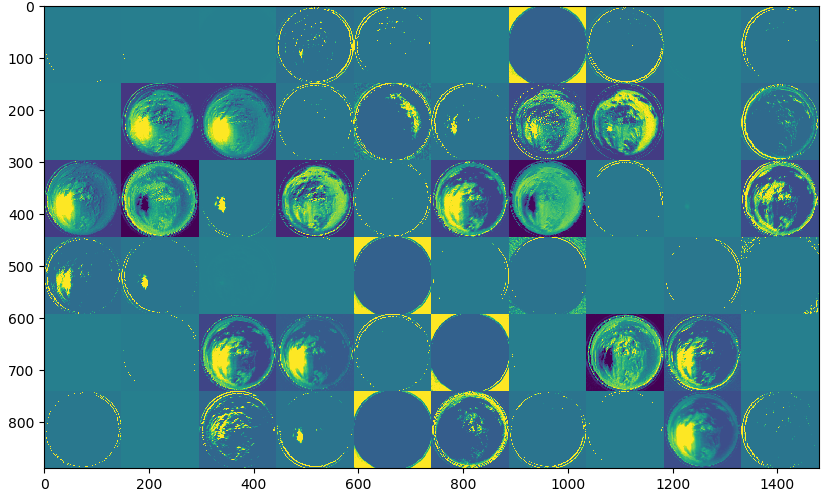}
\caption{2D output resulting from the convolution stage by filters of the first layer}
\label{activation_visualisation}
\end{figure}

\vspace{-0.5\baselineskip}
This leads to the second use of this analysis technique, which is selecting the number of filters per layer in the network. Figure~\ref{activation_visualisation} indicates that the model did not make use of all 60 filters during the training. This gives a hint that decreasing the number of filters in this layer would not penalise the learning process, and would reduce the complexity of the model, i.e shorten the training time while preventing the model from overfitting. In the final configuration, the number of filters per convolutional layer was reduced to 32.

\vspace{-2\baselineskip}

\subsection{Filter Visualisation}

As explained in \cite{dl_with_python}, the different patterns learnt by a model can be visualised by generating, for each filter of the network, the visual pattern that it responds to. To quantify the response of a filter to an image, one can define a loss function returning the average value of the output resulting from the convolution by the given filter. Starting from a noisy image, the visual pattern is then formed using gradient ascent on the pixel of the input to maximise the loss function measuring the response of the filter to that input.

\begin{figure}[H]
\centering
\begin{subfigure}{0.15\textwidth}
    \centering
    \includegraphics[width=1\linewidth]{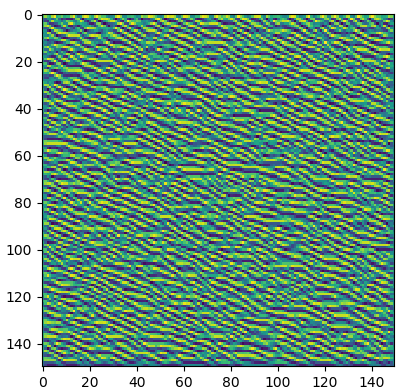}
    %\caption{Long exposure (source:~\cite{sirta})}
    %\label{fig:sub2}
\end{subfigure}
\begin{subfigure}{0.15\textwidth}
    \centering
    \includegraphics[width=1\linewidth]{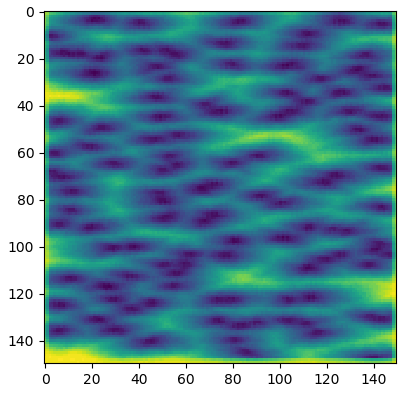}
    %\caption{Long exposure (source:~\cite{sirta})}
    %\label{fig:sub2}
\end{subfigure}
\begin{subfigure}{0.15\textwidth}
    \centering
    \includegraphics[width=1\linewidth]{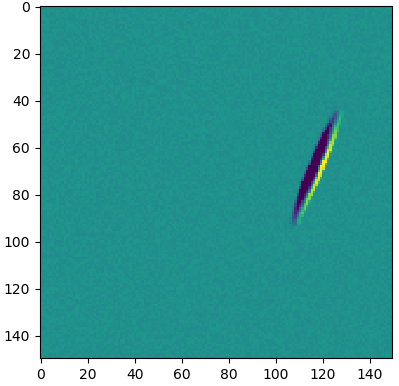}
    %\caption{Short exposure (source:~\cite{sirta})}
    %\label{fig:sub1}
\end{subfigure}%

\begin{subfigure}{0.15\textwidth}
    \centering
    \includegraphics[width=1\linewidth]{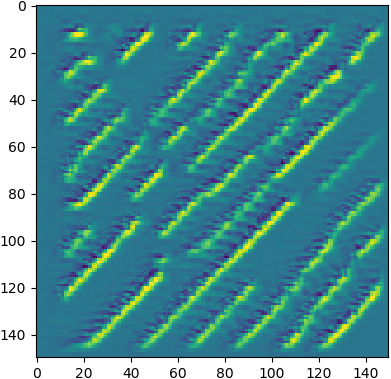}
    %\caption{Long exposure (source:~\cite{sirta})}
    %\label{fig:sub2}
\end{subfigure}
\begin{subfigure}{0.15\textwidth}
    \centering
    \includegraphics[width=1\linewidth]{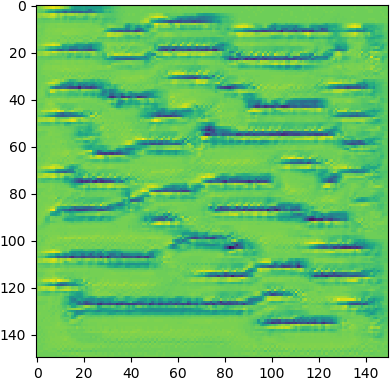}
    %\caption{Long exposure (source:~\cite{sirta})}
    %\label{fig:sub2}
\end{subfigure}
\begin{subfigure}{0.15\textwidth}
    \centering
    \includegraphics[width=1\linewidth]{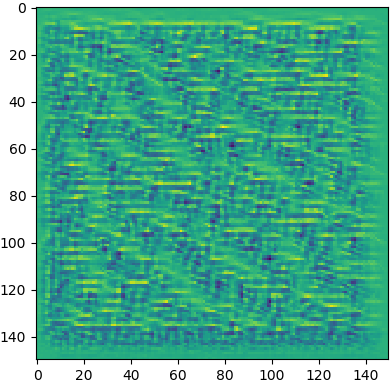}
    %\caption{Short exposure (source:~\cite{sirta})}
    %\label{fig:sub1}
\end{subfigure}%

\begin{subfigure}{0.15\textwidth}
    \centering
    \includegraphics[width=1\linewidth]{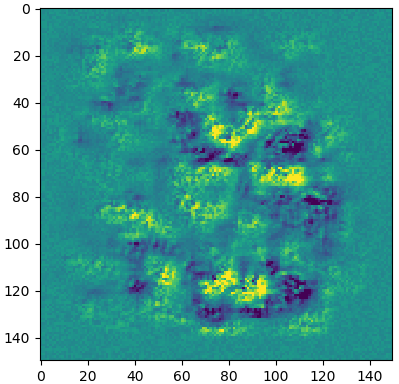}
    %\caption{Long exposure (source:~\cite{sirta})}
    %\label{fig:sub2}
\end{subfigure}
\begin{subfigure}{0.15\textwidth}
    \centering
    \includegraphics[width=1\linewidth]{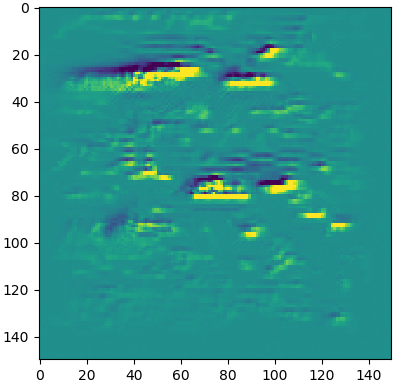}
    %\caption{Long exposure (source:~\cite{sirta})}
    %\label{fig:sub2}
\end{subfigure}
\begin{subfigure}{0.15\textwidth}
    \centering
    \includegraphics[width=1\linewidth]{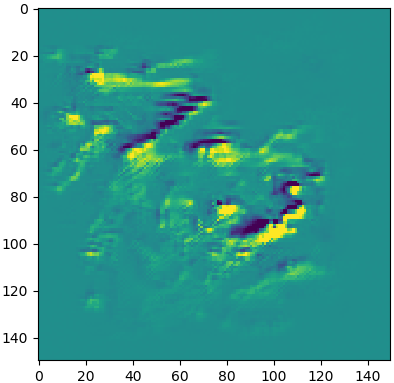}
    %\caption{Short exposure (source:~\cite{sirta})}
    %\label{fig:sub1}
\end{subfigure}%

\caption{Tuned images leading to strong responses by filters of the first, third and fifth convolutional layers (first, second and third row respectively)}
\label{fig:filters_first_layer}
\end{figure}

\vspace{0.5\baselineskip}
Some images generated by this method for filters of the first, third and fifth convolutional layers are presented in Figure~\ref{fig:filters_first_layer}. One can notice that patterns spotted by filters of the first layers are rather abstract and difficult to interpret. However, as we go deeper into the network, corresponding filters respond to patterns of clouds. And the higher the convolutional layer, the more complex the corresponding images: a repetition of small patterns for images of the first and third layers, but complex shapes of cloud covers for the fifth layer. In particular, in Figure~\ref{fig:filters_first_layer}, some filters seem to focus on sparse cloud covers and others at more dense covering.

\vspace{-0.5\baselineskip}

\vspace{-2\baselineskip}

\section{Conclusions}

This study shows that the Deep Learning framework is a promising approach to irradiance forecasting with CNNs being able to extract relevant features from sky images. Also, integrating same day historical data in model predictions proves to be a sensible aspect of short-term forecasting.

%Accurate forecasting of the irradiance and the prediction of sudden changes (ramps) are critical for the integration of solar facilities in the electrical grid. Besides statistical models, the current research in short-term forecast involves the use of hemispherical cameras to capture real time cloud cover changes, which explain the variability of electricity production from solar panels. This study shows that Deep Learning applied to ground based sky imagery is a promising approach to solar energy forecasting. In particular, Deep Learning models fine-tuned on previous samples of the same day strongly improves the overall performance with a MSE forecasting skill based on the smart persistence model reaching 0.44.

\begin{acknowledgements}
%      Part of this work was supported by the German
%      \emph{Deut\-sche For\-schungs\-ge\-mein\-schaft, DFG\/} project
%      number Ts~17/2--1.

The authors would like to acknowledge SIRTA for providing the sky images and irradiance measurements used in this study. We also would like to thank Prof. Philippe Blanc for his valuable advice. This research was supported by Engie, EPSRC and the University of Cambridge.

\end{acknowledgements}


\vspace{-2\baselineskip}
%\addbibresource{library.bib}

%\bibliographystyle{aa}
\bibliography{library.bib}

\end{document}